\pdfoutput=1

\documentclass[11pt]{article}

\usepackage[final]{ACL2023}
\usepackage{times}
\usepackage{latexsym}

\usepackage[T1]{fontenc}
\usepackage[utf8]{inputenc}

\usepackage{microtype}

\usepackage{inconsolata}

\usepackage{comment}
\usepackage{graphicx}
\usepackage{amssymb} % For \cmark
\usepackage{svg}
\usepackage{subcaption}
\usepackage{amsmath}
\usepackage{booktabs}
\usepackage{tabularx}
\usepackage{siunitx} % Required for alignment
\usepackage{subscript}
\usepackage{pifont}
\usepackage{xcolor} % Add to your preamble
\usepackage{tikz}
\usepackage{multirow}
\usepackage{float}  % 引入float包
\usepackage{colortbl}
\usepackage{enumitem} % 提供更多控制列表的功能
\usepackage{latexsym}
\usepackage{amssymb}
\newcolumntype{C}[1]{>{\centering\arraybackslash}p{#1}}

\newcommand{\cmark}{\textcolor{gray}{\ding{51}}} % checkmark
\newcommand{\xmark}{\textcolor{red}{\ding{55}}} % cross

\definecolor{deepgreen}{rgb}{0.0, 0.5, 0.0} 
\newcommand{\correctmark}{\textcolor{deepgreen}{\ding{51}}} % checkmark
\newcommand{\wrongmark}{\textcolor{red}{\ding{55}}} % cross

\title{Improving In-Context Learning with Prediction Feedback for Sentiment Analysis}

\author{Hongling Xu$^{1,3}$, Qianlong Wang$^{1,3}$, Yice Zhang$^{1,3}$, Min Yang$^{4}$
\\ \bf Xi Zeng$^{6}$, Bing Qin$^{5}$, Ruifeng Xu$^{1,2,3}$\thanks{\quad Corresponding Author}
\\
    $^{1}$ Harbin Institute of Technology, Shenzhen, China \\
    $^{2}$ Peng Cheng Laboratory, Shenzhen, China \\
    $^{3}$ Guangdong Provincial Key Laboratory of Novel Security Intelligence Technologies \\
    $^{4}$ SIAT, Chinese Academy of Science ~
    $^{5}$ Harbin Institute of Technology \\
    $^{6}$ The 30th Research Institute of China Electronics Technology Group Corporation \\
     \texttt{xuhongling@stu.hit.edu.cn},
    ~\texttt{xuruifeng@hit.edu.cn} \\
}

\begin{document}
\maketitle
\begin{abstract}

Large language models (LLMs) have achieved promising results in sentiment analysis through the in-context learning (ICL) paradigm. However, their ability to distinguish subtle sentiments still remains a challenge. Inspired by the human ability to adjust understanding via feedback, this paper enhances ICL by incorporating prior predictions and feedback, aiming to rectify sentiment misinterpretation of LLMs. Specifically, the proposed framework consists of three steps: (1) acquiring prior predictions of LLMs, (2) devising predictive feedback based on correctness, and (3) leveraging a feedback-driven prompt to refine sentiment understanding. Experimental results across nine sentiment analysis datasets demonstrate the superiority of our framework over conventional ICL methods, with an average F1 improvement of 5.95\%.\footnote{The source code for our framework is available at \url{https://github.com/HITSZ-HLT/Feedback-ICL}.}

\end{abstract}

% intro 参考文献

\section{Introduction}
Sentiment analysis aims to detect subjective opinions within texts automatically~\cite{medhat2014sentiment}, covering tasks such as sentiment classification, aspect-based sentiment analysis, and emotion detection~\cite{zhang2018deep}. 

Previous studies proposed many supervised methods for sentiment analysis~\cite{xu-etal-2019-bert, li2021dual}. To avoid their reliance on large amounts of human-annotated data, some studies attempted to use limited data to recognize sentiment yet obtained mediocre results. With the advent of large language models (LLMs)~\cite{brown2020language, touvron2023llama}, studies have revealed that LLMs can yield promising performance on sentiment analysis via in-context learning (ICL) paradigm~\cite{li-etal-2023-unified,wang2023chatgpt}, which utilizes only few-shot input-label pairs selected from a candidate example pool.

\begin{figure}[t]
  \centering
  \footnotesize
  \begin{subfigure}[b]{0.425\linewidth}
    \includegraphics[width=\linewidth]{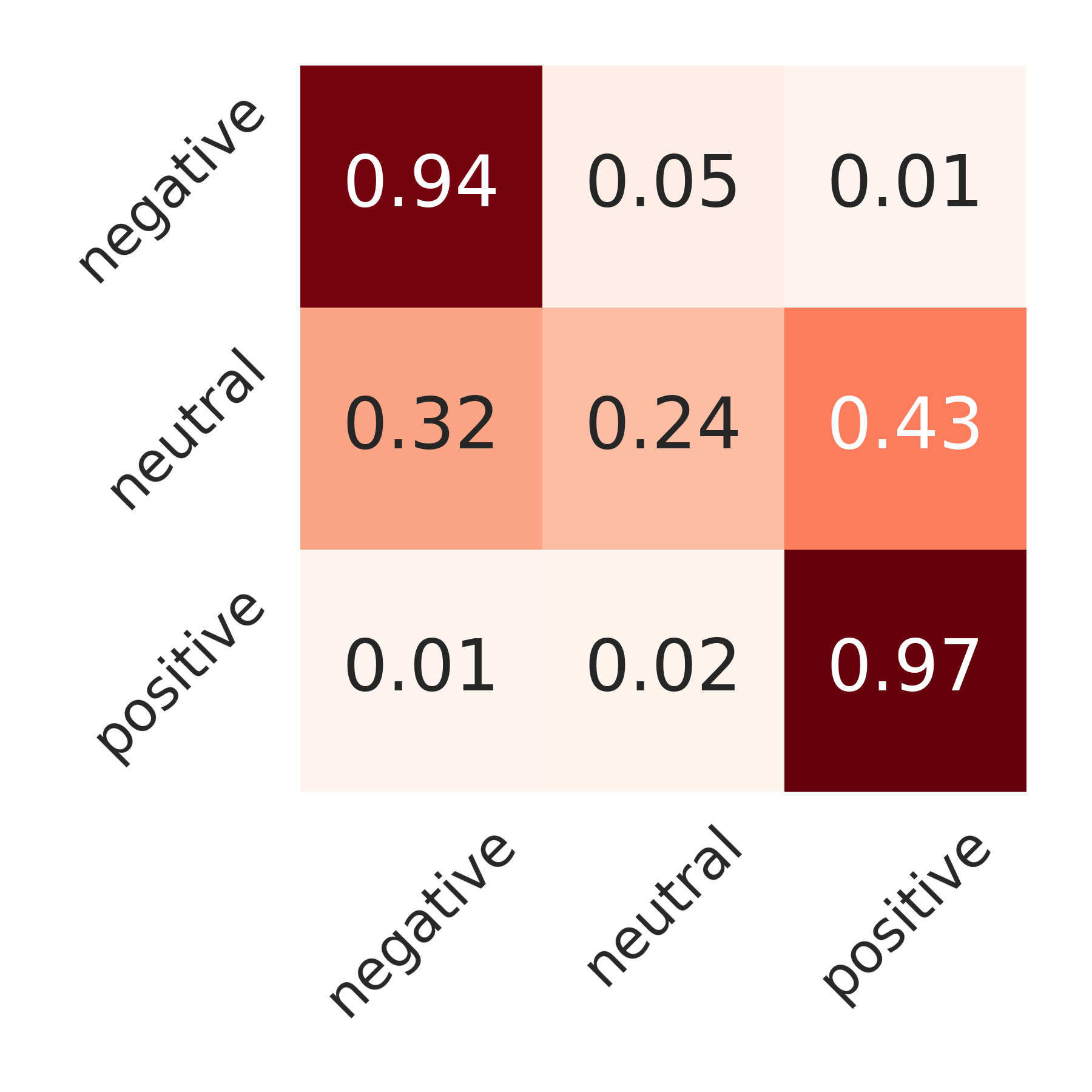}
    \caption{~~~Rest dataset.}
    \label{example:conf_matrix_a}
  \end{subfigure}
  \hfill
  \begin{subfigure}[b]{0.425\linewidth}
    \includegraphics[width=\linewidth]{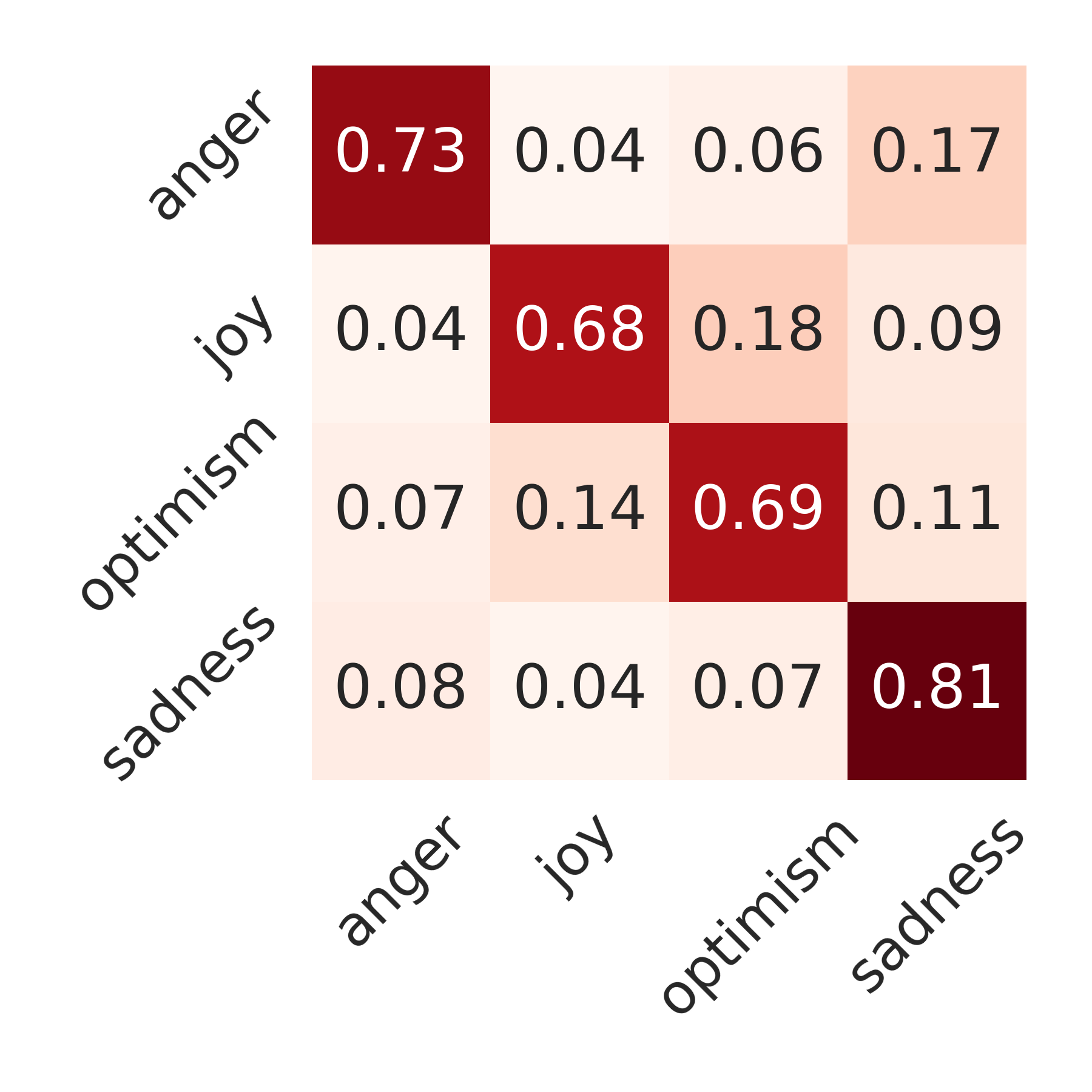}
    \caption{~~~TwEmo dataset.}
    \label{example:conf_matrix_b}
  \end{subfigure}  
  \caption{%Class-wise prediction distribution 
   Normalized confusion matrices on two sentiment analysis datasets. Results are from ChatGPT. }
  \label{fig: example}
\end{figure}

Despite achieving favorable results, the conventional ICL paradigm still faces a concerning limitation. Namely, through the provided examples, LLMs fail to differentiate subtly similar sentiments effectively. Consequently, they would predict plausible yet incorrect sentiment labels. 
As depicted in Figure~\ref{example:conf_matrix_a}, although LLMs can clearly distinguish between \textit{positive} and \textit{negative} polarities, they often mistakenly categorize \textit{neutral} into others.
% even though the given examples include both labels. 
In addition, as shown in Figure \ref{example:conf_matrix_b}, LLMs frequently mislabel fine-grained sentiments as relevant but wrong labels, such as \textit{joy} and \textit{optimism}, stemming from their incapacity to understand nuanced sentiments with similar contexts.

% but have difficulty empathizing with fine-grained emotions such as anger and sadness. This issue also occurs with neutral sentiments, where LLMs often mistakenly categorize \textit{neutral} into \textit{negative} or \textit{positive}.

\begin{comment}
\begin{figure}[tbp]
    \centering
    \includegraphics[width=0.85\linewidth]{figs/poem_prob.pdf}
    \caption{Different ICL Methods. Examples include input-label pairs and PF denotes Prediction Feedback.}
    \label{fig:enter-label}
\end{figure}
\end{comment}

% When solving problems, humans generally first make plans (or predictions) based on prior knowledge and then adjust their understanding and thinking based on the actual feedback from answers \cite{belanger2011theories}. Such a process enables humans to better understand and distinguish similar concepts. By contrast, in the ICL paradigm, LLMs can only see the correct sentiment labels and fail to receive feedback to adjust their understanding and reasoning about the input-label pairs. Inspired by it, we assume that if demonstrations provide valuable feedback on prior predictions, LLMs would elicit a self-adjustment process akin to human learning, which may correct some understanding mistakes. 

\begin{figure*}[t]
    \centering
    \includegraphics[width=0.95\textwidth]{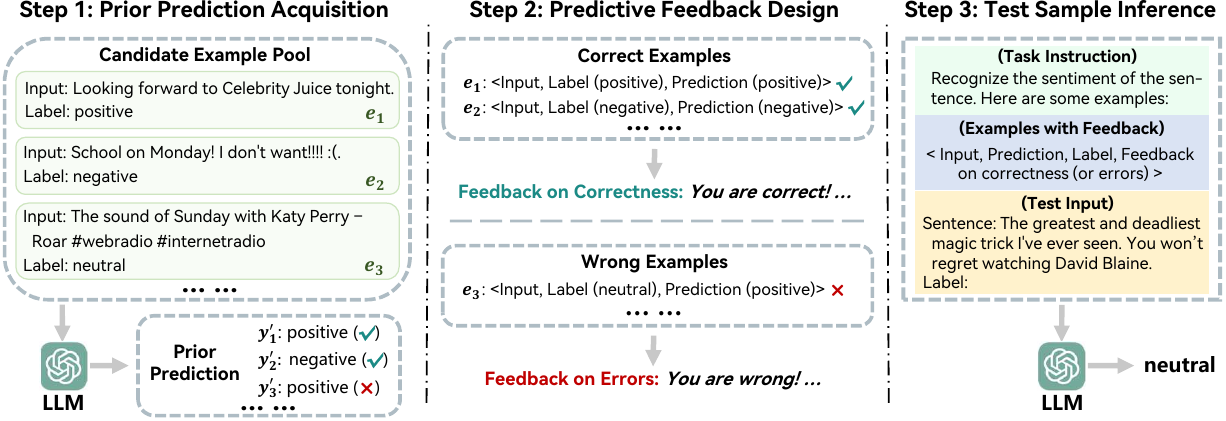}
    \caption{Overview of our framework.}
    \label{fig:overview}
\end{figure*}

Inspired by the human learning process, where individuals initially make plans based on prior knowledge and adjust their understanding through actual feedback~\cite{belanger2011theories}, 
we propose to integrate feedback on prior predictions into ICL, aiming to rectify sentiment misunderstandings of LLMs.
Specifically, our framework first yields prior predictions for each candidate example using traditional ICL. 
We then categorize examples into two sub-pools based on correctness and exploit feedback to illustrate differences between prior predictions and human annotations. Finally, during inferring, we select relevant examples from each sub-pool and use a specific feedback-driven prompt to wrap input, prediction, label, and feedback. 
Unlike conventional ICL, where LLMs only see correct labels, our framework effectively directs LLMs to adjust their sentiment understanding and reasoning to align more closely with label perception through prediction and feedback.

Experimental results on nine sentiment analysis datasets show that our framework outperforms existing ICL baselines by 5.95\% in average F1. Further discussions indicate its effectiveness and robustness. Moreover, the framework also yields competitive results when extended to other tasks.

\section{Preliminary}
Sentiment analysis aims to predict the sentiment label $y'$ of an input text $x$. Here, different tasks may have different label spaces $\mathcal{C}$ and inputs.\footnote{For example, aspect sentiment classification task needs to consider the effect of aspects in inputs \cite{pontiki-etal-2014-semeval}.}
In ICL paradigm, given an input $x$ and $k$-shot in-context examples $\{(x_i, y_i)\}^k_{i=1}$ retrieved from a pre-defined candidate pool $\mathcal{P}$ (its size is relatively small), a frozen LLM $\mathcal{M}$ is used to predict $y'$. 
\begin{equation}
y' = \operatorname*{argmax}_{y\in \mathcal{C}} \mathcal{M}(y|(x_1,y_1),...(x_k,y_k), x)
\end{equation}
where we ignore the task instruction and example template for simplicity. To avoid irrelevant outputs, we employ a constrained decoding strategy that ensures only label words within $\mathcal{C}$ can be generated. Besides, we directly use these label words as the verbalizer for each class in classification.\footnote{If a label word is split into subtokens, we use only the first subword for prediction, such as 'optim' for 'optimism'.}

\section{The Proposed Framework}
As shown in Figure~\ref{fig:overview}, our framework consists of three steps:
1) \textit{prior prediction acquisition},
2) \textit{predictive feedback design}, and
3) \textit{test sample inference}. Below is a detailed description. % of each step.

\paragraph{Step 1: Prior Prediction Acquisition.}
This step focuses on acquiring the prior prediction $y'_i$ on each candidate example $x_i$ for subsequent feedback provision. To this end, examples from $\mathcal{P}$ are treated as inference targets. 
Following the traditional ICL, we randomly select other four input-label pairs from the candidate pool as demonstrations,\footnote{The reason for selecting four is to strike a trade-off between contextual richness and computational efficiency.} which are combined with the task instruction to prompt the LLM for predictions (see Appendix \ref{sec: prompt} for more information). We refer to these predictions as \textit{prior predictions} because they serve to reflect the prior sentiment understanding of LLMs.

\begin{table*}[tbp]
  \centering
  \footnotesize
    \begin{tabular} {p{1.2cm}C{1.14cm}C{1.14cm}C{1.14cm}C{1.14cm}C{1.14cm}C{1.14cm}C{1.14cm}C{1.14cm}C{1.14cm}}
    \toprule
    \multirow{2}{*}{Method} & \multicolumn{4}{c}{\textbf{\footnotesize{Sentiment Classification}}} & \multicolumn{3}{c}{\textbf{\footnotesize{Aspect Sentiment Classification}}} & \multicolumn{2}{c}{\textbf{\footnotesize{Emotion Detection}}} \\
    \cmidrule(lr){2-5} \cmidrule(lr){6-8} \cmidrule(lr){9-10}
    & \multicolumn{1}{c}{\texttt{SST-2}} & \multicolumn{1}{c}{\texttt{TwSenti}} & \multicolumn{1}{c}{\texttt{Poem}} & \multicolumn{1}{c}{\texttt{Finance}} & \multicolumn{1}{c}{\texttt{Rest}} & \multicolumn{1}{c}{\texttt{Laptop}} & \multicolumn{1}{c}{\texttt{Twitter}} & \multicolumn{1}{c}{\texttt{EmoC}} & \multicolumn{1}{c}{\texttt{TwEmo}} \\ 
    \midrule
     BERT\scalebox{1.0}[1.0]{-}FT\textsuperscript{\textdagger} & 84.69 & 54.54 & 72.55 & 89.41 & 64.59 & 69.03 & 56.40 & 47.73 & 62.53 \\
    
    \midrule
    Random & 89.82 & 55.27 & 55.08 & 75.34  & 68.77 & 73.02 & 54.95 & 45.20 & 47.77\\
    ~~~+~Ours & \cellcolor[gray]{0.9} 91.65\textsubscript{+1.83} & \cellcolor[gray]{0.9} 60.33\textsubscript{+5.06} & \cellcolor[gray]{0.9}{64.37\textsubscript{+9.29}} & \cellcolor[gray]{0.9}{78.64\textsubscript{+3.30}} & \cellcolor[gray]{0.9}{71.16\textsubscript{+2.39}} & 72.80\textsubscript{-0.22} & \cellcolor[gray]{0.9}{57.64\textsubscript{+2.69}} & \cellcolor[gray]{0.9}{52.50\textsubscript{+7.30}} & \cellcolor[gray]{0.9}{60.91\textsubscript{+13.14}} \\
    \midrule
    
    BM25 & 90.26 & 55.35 & 49.99 & 56.13 & 68.99 & 70.29 & 50.99 & 44.89 & 48.44 \\
    ~~~+~Ours & \cellcolor[gray]{0.9}{91.85\textsubscript{+1.59}} & \cellcolor[gray]{0.9}{59.20\textsubscript{+3.85}} & \cellcolor[gray]{0.9}{61.27\textsubscript{+11.28}} & \cellcolor[gray]{0.9}{66.94\textsubscript{+10.81}} & \cellcolor[gray]{0.9}{71.76\textsubscript{+2.77}} & \cellcolor[gray]{0.9}{71.67\textsubscript{+1.38}} & \cellcolor[gray]{0.9}{56.22\textsubscript{+5.23}} & \cellcolor[gray]{0.9}{51.63\textsubscript{+6.73}} & \cellcolor[gray]{0.9}{62.88\textsubscript{+14.44}} \\
    \midrule
    
    SBERT & 87.96 & 50.13 & 47.41 & 47.12 & 68.21 & 65.72 & 50.60  & 46.28 & 48.58 \\
    ~~~+~Ours & \cellcolor[gray]{0.9}{91.57\textsubscript{+3.61}} & \cellcolor[gray]{0.9}{55.08\textsubscript{+4.95}} & \cellcolor[gray]{0.9}{56.42\textsubscript{+9.01}} & \cellcolor[gray]{0.9}{58.21\textsubscript{+11.09}} & \cellcolor[gray]{0.9}{71.29\textsubscript{+3.08}} & \cellcolor[gray]{0.9}{69.56\textsubscript{+3.84}} & \cellcolor[gray]{0.9}{56.07\textsubscript{+5.47}}  & \cellcolor[gray]{0.9}{50.30\textsubscript{+4.02}} & \cellcolor[gray]{0.9}{61.22\textsubscript{+12.64}} \\
    \midrule
    
    MMR & 89.64 & 50.80 & 49.74 & 54.51 & 68.30 & 66.72 & 51.07 & 43.72 & 49.94 \\
    ~~~+~Ours & \cellcolor[gray]{0.9}{92.65\textsubscript{+3.01}} & \cellcolor[gray]{0.9}{56.84\textsubscript{+6.04}} & \cellcolor[gray]{0.9}{63.38\textsubscript{+13.64}} & \cellcolor[gray]{0.9}{59.85\textsubscript{+5.34}} & \cellcolor[gray]{0.9}{69.76\textsubscript{+1.46}} & \cellcolor[gray]{0.9}{69.23\textsubscript{+2.51}} & \cellcolor[gray]{0.9}{55.57\textsubscript{+4.50}} & \cellcolor[gray]{0.9}{49.31\textsubscript{+5.59}} & \cellcolor[gray]{0.9}{61.74\textsubscript{+11.80}} \\
    \midrule
    
    K-Means & 88.74 & 56.26 & 51.39 & 76.14 & 71.01 & 73.68 & 55.20 & 45.71 & 46.93 \\
    ~~~+~Ours & \cellcolor[gray]{0.9}{92.23\textsubscript{+3.49}} & \cellcolor[gray]{0.9}{61.32\textsubscript{+5.06}} & \cellcolor[gray]{0.9}{68.70\textsubscript{+17.31}} & \cellcolor[gray]{0.9}{78.44\textsubscript{+2.30}} & \cellcolor[gray]{0.9}{71.10\textsubscript{+0.09}} & 73.11\textsubscript{-0.57} & \cellcolor[gray]{0.9}{57.78\textsubscript{+2.58}} & \cellcolor[gray]{0.9}{53.72\textsubscript{+8.01}} & \cellcolor[gray]{0.9}{61.89\textsubscript{+14.96}} \\
    \bottomrule
  \end{tabular}
  \caption{Main results in F1\% (see Acc\% results in Appendix~\ref{sec: more_acc}). Fine-tuning methods are marked by \textdagger.} 
  \label{tab: main}
\end{table*}

\paragraph{Step 2: Predictive Feedback Design.}
The correctness of the prior predictions directly indicates whether LLMs can accurately grasp the sentiment of the corresponding examples. 
To elicit self-adjustments of LLMs in understanding and reasoning, we first classify the examples into two sub-pools, $\mathcal{P}_{c}$ and $\mathcal{P}_w$, where the former includes correctly classified examples, and the latter contains wrong ones. 
We then provide each sub-pool with feedback in the natural language form:

\textbf{feedback on $\mathcal{P}_{c}$}: \textit{You are correct! Stay determined and keep moving forward.}

\textbf{feedback on $\mathcal{P}_w$}: \textit{You are wrong! Make sure your prediction is accurate.}

\paragraph{Step 3: Test Sample Inference.}
To complete the inference for the given test input, we first retrieve $k/2$ examples from each candidate sub-pool. Since our framework is retrieval-mode agnostic, any example retrieval technique can be employed here. In addition, we develop a feedback-driven prompt template to wrap the input, prediction, label, and feedback of each selected example into a quadruple. Subsequently, these quadruples are organized by $\mathcal{P}_w$ examples before $\mathcal{P}_{c}$ ones and sorted by descending relevance. 
Finally, the test input is set in the standard example template, with the label position left blank for prediction. 

\begin{comment}
    \begin{equation}
    \begin{split}
        y'_{i} = \text{verbalizer}(\mathcal{M}(y|TI, Demo', T(x_i))) \\
        Demo' = \{ T'(x_i, y'_i, y_i, fd) \mid i = 1,..., k\}
    \end{split}
\end{equation}
\end{comment}

\section{Experiments}

% Random  & 86.41\textsubscript{+0.00} & 75.83\textsubscript{+0.00} & 85.44\textsubscript{+0.00} & 80.49\textsubscript{+0.00} \\
%    ~+ Ours & 87.09\textsubscript{+0.00} & 79.85\textsubscript{+0.00} & 85.15\textsubscript{+0.00} & 80.29\textsubscript{+0.00} \\

\subsection{Experimental Setup}
\paragraph{Datasets.} We conduct experiments across three sentiment analysis tasks using nine distinct datasets, including 
\textbf{Sentiment Classification} (SC): SST-2~\cite{socher-etal-2013-recursive}, TwSenti~\cite{rosenthal-etal-2017-semeval}, Poem~\cite{sheng-uthus-2020-investigating}, and Finance~\cite{malo2014good}; \textbf{Aspect Sentiment Classification} (ASC): Rest and Laptop~\cite{pontiki-etal-2014-semeval}, and Twitter~\cite{dong-etal-2014-adaptive}; \textbf{Emotion Detection} (ED): EmoC~\cite{chatterjee-etal-2019-semeval} and TwEmo~\cite{barbieri-etal-2020-tweeteval}. Detailed statistics are listed in Appendix~\ref{sec: dataset}.

\paragraph{Baselines.}
To evaluate the effectiveness of the proposed framework, we combine it with various training-free example retrieval baselines for comparison, including \textbf{Random}, \textbf{BM25}~\cite{robertson2009probabilistic}, \textbf{SBERT}~\cite{reimers-gurevych-2019-sentence}, \textbf{MMR}~\cite{ye-etal-2023-complementary}, and
\textbf{K-Means}~\cite{zhang2023automatic}. Furthermore, we introduce \textbf{BERT-FT}, where the BERT-base model is fine-tuned directly on candidate pool examples.
See Appendix~\ref{baseline} for their specific settings.

\paragraph{Implementation Details.}
In this study, we utilize Llama-2 13B Chat \cite{touvron2023llama2} as the backbone LLM due to its moderate scale and excellent ICL performance. Unless otherwise mentioned, we set the number of in-context examples to 4. 
The candidate pool is formed by sampling 300 label-balanced examples from each training set. We experiment with 3 different random seeds and present the average results. More implementation details are shown in Appendix~\ref{sec: settings}.

\subsection{Main Results}

% 第一段重点讲提升 
Results shown in Table~\ref{tab: main} indicate that our framework substantially enhances baseline performance on nearly all datasets. For example, augmenting K-means with our framework results in an average F1 increase of 5.91\%, exhibiting its superiority.  
Meanwhile, compared with BERT-FT, our approach demonstrates outstanding performance on the majority of datasets, highlighting its efficacy in resource-limited scenarios. 

Additionally, our framework notably excels in ED, where detecting subtly similar sentiments is crucial. Comparatively, ASC involves more complex aspect-based contextual understanding, constraining the improvement in these tasks. 

Contrary to previous studies~\cite{rubin-etal-2022-learning, li-etal-2023-unified}, we find that semantic similarity retrievals like SBERT negatively impact the performance. We suppose it is due to the demonstration bias when solving simple sentiment analysis tasks~\cite{fan2023comparable} and the lack of example complementarity~\cite{ye-etal-2023-complementary}.

\begin{table}[t]
  \centering
  \footnotesize
    \begin{tabular}{@{}cc|cc|ccc@{}}
    \toprule
    % \multicolumn{2}{c|}{ICL Input} & \multicolumn{2}{c|}{Ours} & \multicolumn{3}{c}{Dataset} \\
    % \midrule
    Inst & Label & Pred & Feed & Poem & Rest & TwEmo \\
    \addlinespace[0.15cm]
    % \midrule
    \cmark & \cmark & \cmark & \cmark & \textbf{68.70} & 71.10 & \textbf{61.89} \\
    \midrule
   % \cmark &  & \cmark & &  &  &  \\
   %  &  & \cmark & &  &  &  \\
    \xmark & \cmark & \cmark & \cmark & 55.97 & \textbf{71.70} & 61.13  \\
    \cmark & \xmark & \cmark & \cmark & 51.47 & 67.05 & 52.78  \\
    \cmark & \cmark & \xmark & \cmark & 59.47 & 69.94 & 48.53 \\
    \cmark & \cmark & {{\textit{R}}} & \cmark & 67.00 & 70.14 & 60.71 \\
    \cmark & \cmark & {{\textit{Z}}} & \cmark & 64.71 & 70.18 & 60.91 \\
    \cmark & \cmark & \cmark & \xmark & 63.46 & 70.49 & 60.02 \\
    \bottomrule
  \end{tabular}
  \caption{Ablation study based on K\scalebox{1.0}[1.0]{-}Means. `Inst' for instructions, `Pred' for predictions, and `Feed' for feedback. We explore additional sources of Pred including random errors (R) and zero-shot prompting (Z).}  % (see more in Appendix~\ref{more_ablation})
  \label{tab:ablation_kmeans}
\end{table}

\subsection{Ablation Study}

We perform the ablation study to explore the effect of each component, as presented in Table~\ref{tab:ablation_kmeans}.
When removing task instructions, we see performance drops except for the Rest dataset, indicating its insensitivity to instructions with information-rich inputs. 
Additionally, both the removal of labels and prior predictions cause a notable decline, by averages of 10.13\% and 7.92\% respectively, highlighting the significance of their combination in our framework. 
Besides, employing alternative prediction sources or excluding feedback also leads to a slight decrease.

\subsection{Effect on Subtle Sentiments} 
\begin{figure}[h]
    \centering
       \includegraphics[width=0.85\linewidth]{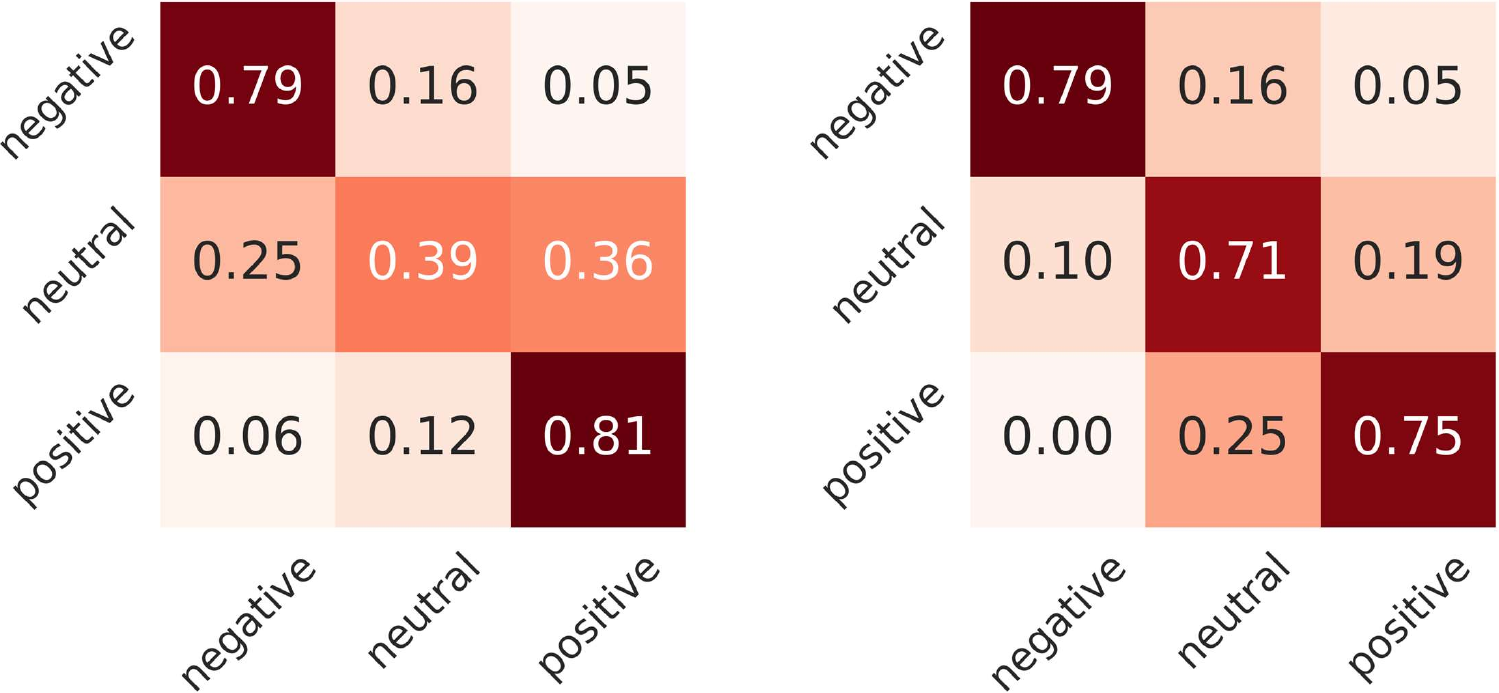}
        \caption{Normalized confusion matrices for the Poem dataset:
K-Means (left) and K-Means+Ours (right). See results for more datasets in Appendix~\ref{sec: more_effect}.}
        \label{fig:conf_poem}
\end{figure}

To demonstrate the impact of our framework on subtle similar sentiments, we visualize the prediction distributions as depicted in Figure~\ref{fig:conf_poem}. We can observe an obvious change of distribution in \textit{neutral}, whose correct rate increases by 32\%, while the other two categories are relatively stable. These results suggest that integrating predictive feedback could make more accurate distinctions between subtly similar sentiments.

\subsection{Discussions\footnote{We present more analyses in Appendix~\ref{sec: appendix more_analysis}, including The Sensitivity of Feedback Prompt, Impact of the Number of Examples, and Impact of the Order of Examples.}}

\begin{table}[!t]
  \centering
  \footnotesize
    \begin{tabular}{C{1.6cm}p{1.3cm}C{1.4cm}C{1.4cm}}
    \toprule
     {LLM} & Method & Rest & TwEmo \\
    \midrule
    \multirow{6}{*}{Mistral 7B} 
    & Random & 73.66 & 66.30 \\
     \multirow{6}{*}{Instruct} &~~+~Ours  & \textbf{75.05\textsubscript{+1.39}} & \textbf{68.50\textsubscript{+2.20}} \\
     \cmidrule{2-4}
    & BM25 & 72.03 & 68.47 \\
    &~~+~Ours  & \textbf{73.47\textsubscript{+1.44}} & \textbf{69.74\textsubscript{+1.27}} \\
    \cmidrule{2-4}
    & K\scalebox{1.0}[1.0]{-}Means & 73.64 & 66.07 \\
    &~~+~Ours   & \textbf{74.08\textsubscript{+0.44}} & \textbf{68.04\textsubscript{+1.97}} \\
    
    \midrule
    
    \multirow{6}{*}{GPT-3.5} 
     & Random & 68.46 & 69.97 \\
     \multirow{6}{*}{Turbo}  &~~+~Ours  & \textbf{77.87\textsubscript{+9.41}} & \textbf{72.11\textsubscript{+2.14}} \\
     \cmidrule{2-4}
    & BM25 & 70.15 & 68.17 \\
    &~~+~Ours  &\textbf{74.34\textsubscript{+4.19}} &\textbf{71.19\textsubscript{+3.02}} \\
    \cmidrule{2-4}
    & K\scalebox{1.0}[1.0]{-}Means & 70.15 & 69.68 \\
    &~~+~Ours &\textbf{77.29\textsubscript{+7.14}} &\textbf{72.12\textsubscript{+2.44}} \\
    
    \midrule
    
    \multirow{2}{*}{GPT-4} & K\scalebox{1.0}[1.0]{-}Means & 81.90 & 77.80 \\
    &~~+~Ours  &\textbf{\textbf{82.29}\textsubscript{+0.39}} &\textbf{\textbf{78.10}\textsubscript{+0.30}} \\
    %\midrule
    
    \bottomrule
  \end{tabular}
  \caption{Results of Different LLMs (F1\%).} 
  \label{tab: lm generalization}
\end{table}

\paragraph{Language Model Generalization.} 
To evaluate the adaptability of our framework, we conduct model generalization experiments with various LLMs. Specifically, we select three capable and prominent models: Mistral 7B Instruct~\cite{jiang2023mistral}, GPT-3.5 Turbo~\cite{ouyang2022training}, and GPT-4~\cite{openai2023gpt4}.\footnote{The exact versions of the three models are as follows: Mistral-7B-Instruct-v0.2, gpt-3.5-turbo-0301, and gpt-4-0613.} As illustrated in Table~\ref{tab: lm generalization}, incorporating our framework consistently enhances the performance of ICL, particularly with GPT-3.5 Turbo, where the average F1 improvement is 4.72\%, illustrating its generalizability.
Furthermore, employing LLMs with more advanced comprehension such as GPT-4 significantly improves sentiment analysis results. 

\begin{table*}[t]
    \centering
    \footnotesize
    \begin{tabular}{p{0.9cm}C{6.85cm}C{6.85cm}}
        \toprule
         & \textbf{Conventional In-context Learning} & \textbf{In-context Learning w/ Prediction Feedback} \\
         
        \cmidrule(lr){2-2} \cmidrule(lr){3-3}
        
        Prompt & \textit{Task Instruction}: \ldots~~\textit{Examples}: $\{(x_i, y_i)\}$ & 
        \textit{Task Instruction}: \ldots~~\textit{Examples}: $\{(x_i, y'_i, y_i, fd)\}$ \\
        \midrule
        % \cmidrule(lr){2-3}
        Input & \multicolumn{2}{p{13.7cm}}{@user Wishing you well sir... you are an extremely straightforward and jovial person...} \\
        \cmidrule(lr){2-3}
        Output & optimism (\wrongmark) & joy (\correctmark) \\
        \cmidrule(lr){2-2} \cmidrule(lr){3-3}
        % {\textcolor{deepgreen}{\ding{51}}} {\textcolor{red}{\ding{55}}}
        \multirow{7}{*}{Reason} & \multicolumn{1}{p{6.85cm}}{ The sentence conveys a positive and uplifting message. The use of the word "jovial" suggests \textbf{\textcolor{red}{a lighthearted and cheerful tone, which is consistent with the label "optimism"}}. Additionally, the phrase "wishing you well" implies good intentions and a \textbf{\textcolor{red}{positive attitude towards the recipient, further supporting the label "optimism".}}} & \multicolumn{1}{p{6.85cm}}{ The sentence is addressed to a specific user and expresses well-wishes, which suggests a positive and friendly tone. The description of the user as "extremely straightforward and jovial" further \textbf{\textcolor{deepgreen}{reinforces the positive sentiment. Based on the examples and feedback provided, the label "joy" best captures the emotion of the sentence.}}}\\
        \midrule
         Input & \multicolumn{2}{p{13.7cm}}{they beached their galley on the shore, and sought} \\
        \cmidrule(lr){2-3}
        Output & positive (\wrongmark) & neutral (\correctmark) \\
        \cmidrule(lr){2-3}
        \multirow{6}{*}{Reason} & \multicolumn{1}{p{6.85cm}}{ The sentence has a sense of adventure and exploration, with the use of the word "beached" which implies \textbf{\textcolor{red}{a successful landing,}} and the word "sought" which implies a \textbf{\textcolor{red}{deliberate and purposeful action. The overall tone of the sentence is optimistic and confident, which aligns with the positive label.}}} & \multicolumn{1}{p{6.85cm}}{ 
        \textbf{\textcolor{deepgreen}{The sentence does not convey a positive or negative sentiment.}} In this sentence, the word "beached" is used in the sense of landing on the shore, which is \textbf{\textcolor{deepgreen}{a neutral action.}} Additionally, the word "sought" \textbf{\textcolor{deepgreen}{does not convey any strong emotions or opinions, and is also a neutral word. Therefore, the overall sentiment of the sentence is neutral.}}}\\ 
        \bottomrule
    \end{tabular}
    \caption{Case study of conventional ICL and our framework. More cases are shown in Appendix~\ref{sec: more_case}.}
    \label{tab: case}
\end{table*}

\begin{figure}[t]
    \centering
    \includegraphics[width=0.99\linewidth]{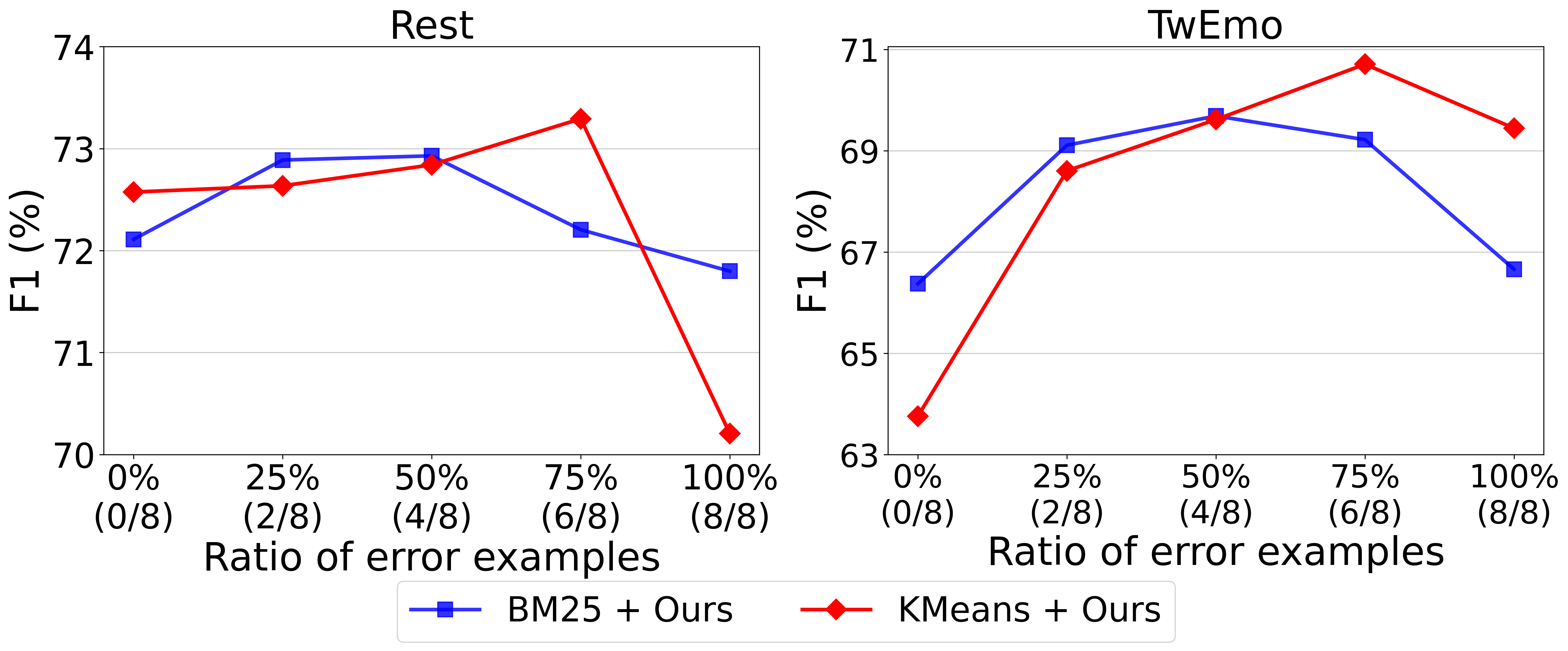}
    \caption{Impact of the error example ratio. }
    \label{fig: ratio}
\end{figure}

\paragraph{Impact of the Ratio of Error Examples.} To assess the effect of the erroneous examples, we fix $k$ at 8 and vary the number of examples selected from $P_w$, where the ratio is adjusted in 1/4 increments, ranging from 0 to 1.
The results are depicted in Figure~\ref{fig: ratio}. We find that the framework underperforms with either no error examples or an excess of them. Meanwhile, as the quantity of incorrect examples increases, the performance initially rises and then declines, indicating that a relative balance of incorrect to correct examples is beneficial.

\paragraph{Task Generalization.}
To demonstrate that our framework is not confined to adjusting sentiment understanding of LLMs, we conduct experiments on three additional datasets: P-Stance~\cite{li-etal-2021-p} for stance detection, TwIrony~\cite{van-hee-etal-2018-semeval} for irony detection, and MNLI~\cite{wang2018glue} for natural language inference (NLI). 
Results are illustrated in Table~\ref{tab: others}.
Notably, the proposed framework also yields significant improvements on these datasets, with average F1 increasing by 1.85\% for P-Stance, 3.88\% for TwIrony, and 5.35\% for MNLI. These results suggest that the adaptability of our prediction feedback mechanism can extend to a broader scope of language understanding tasks. 

\begin{table}[t]
  \centering
  \footnotesize
    \begin{tabular}{p{1.2cm}C{1.5cm}C{1.5cm}C{1.5cm}}
    \toprule
     & P-Stance & TwIrony & MNLI \\
    \midrule
    Random     & 70.94 & 62.29 & 49.63  \\
    ~~+~Ours   & \textbf{73.31\textsubscript{+2.37}} & \textbf{65.44\textsubscript{+3.15}} &  \textbf{55.21\textsubscript{+5.58}}  \\
    \midrule
    BM25       & 72.23 & 60.06 & 50.68  \\
    ~~+~Ours   & \textbf{72.98\textsubscript{+0.75}} & \textbf{64.29\textsubscript{+4.23}} & \textbf{56.65\textsubscript{+5.97}} \\
    \midrule
    K-Means    & 71.17 & 61.47 & 50.60  \\
    ~~+~Ours   & \textbf{\textbf{73.60}\textsubscript{+2.43}} & \textbf{\textbf{65.72}\textsubscript{+4.25}} & \textbf{55.09\textsubscript{+4.49}}  \\
    % Add more rows as needed 
    \bottomrule
  \end{tabular}
  \caption{Results of task generalization (F1\%).}
  \label{tab: others}
\end{table}

\paragraph{Case Study.}
To gain a deeper insight into the advantages of our framework, we perform the case study focusing on the outputs and explanations of the LLM, as detailed in Table~\ref{tab: case}. 
In Case 1, our framework identifies the emotion as \textit{joy} rather than the plausible yet incorrect \textit{optimism}, and offers a more fitting explanation that aligns with the implied emotion. In Case 2, our framework correctly identifies the \textit{neutral} polarity and avoids inaccurate sentiment interpretation. 
These cases reveal that our framework promotes self-adjustment of the LLM in sentiment analysis, refining both output and reasoning accuracy.

\section{Conclusion}
In this paper, we propose a novel ICL framework that utilizes prediction feedback akin to human learning. It improves ICL by incorporating both prior predictions and corresponding feedback into examples, tackling the difficulties LLMs encounter when identifying subtle sentiments. Experiments across various datasets confirm the advantages of our framework compared to conventional ICL, as well as its potential for broader applications.

\section*{Acknowledgments} 
This work was partially supported by the National Natural Science Foundation of China 62176076,  Natural Science Foundation of GuangDong 2023A1515012922, the Shenzhen Foundational Research Funding JCYJ20220818102415032 (JCYJ20210324115614039), the Major Key Project of PCL2021A06, Guangdong Provincial Key Laboratory of Novel Security Intelligence Technologies 2022B1212010005.

\section*{Limitations}

While our research significantly enhances the performance of conventional ICL and provides in-depth analyses about the adjustment for sentiment understanding of LLMs, the inner mechanisms of the framework remain elusive due to the black-box nature of language models. 
Besides, our research primarily focuses on sentiment analysis and text classification within NLU, leaving more complex areas like text summarization and commonsense generation unexplored. We aim to broaden the scope of our framework in future work, exploring more insights and wider applicability.

\bibliography{custom}
\bibliographystyle{acl_natbib}

\clearpage
\newpage

\appendix

\section*{\centering\textbf{Appendix for ``Improving In-Context Learning with Prediction Feedback for Sentiment Analysis''}}

We organize the appendix into four sections:

\begin{itemize}[leftmargin=*,nosep]
    \item  Prompts used in the proposed framework are presented in Appendix~\ref{sec: prompt};
    \item Additional details of datasets, baselines, and implementation are presented in Appendix~\ref{sec: detail};
    \item Additional experimental results in different settings, such as more metrics, baselines, and datasets are presented in Appendix~\ref{sec: appendix more_results}; and
    \item More discussions about the proposed framework are presented in Appendix~\ref{sec: appendix more_analysis}.
\end{itemize}

\section{Prompt Design}
\label{sec: prompt}

We present the task instructions and prompt templates utilized in our framework for each task in Table~\ref{tab: prompt}. Besides, for conventional ICL, the examples and test input are wrapped in the template that removes prediction and feedback, and the formatting word "Correct Label:" is replaced by "Label:".

We divide the feedback prompt into two parts, namely, feedback on correctness and feedback for analysis, abbreviated as FC and FA. Two manually designed feedback prompts are illustrated in Table~\ref{tab: feedback_prompt} for further discussion (see~\ref{sec: sensitivity}). 

To generate explanations in the case study, we construct instructive forms\footnote{\url{https://github.com/huggingface/blog/blob/main/llama2.md\#how-to-prompt-llama-2}} and employ prompts: \textit{Provide the correct label for the following sample and explain your answer based on the above examples (and feedback).}

\section{Detailed Settings of Experiments}
\label{sec: detail}

\subsection{Dataset and Metrics}
\label{sec: dataset}
We provide detailed statistics of each investigated dataset in Table~\ref{tab: dataset}. When establishing the candidate pool, we select instances only from the training set. Additionally, for Finance, lacking a standard split, we randomly allocate 20\% of total samples to both the dev and test sets. For MNLI that does not offer publicly available test set labels, we evaluate by the dev set. 

Across all sentiment analysis datasets utilized in this study, we uniformly apply two metrics for evaluation: Accuracy (\textbf{Acc}) and F1 score (\textbf{F1}). Moreover, we use binary-F1 for binary classification tasks and macro-F1 for all others.

\subsection{Baseline Details}
\label{baseline}
(1) \textbf{Random} randomly selects $k$-shot examples from the candidate pool for each test sample. 

\noindent(2) \textbf{BM25}~\cite{robertson2009probabilistic} assesses relevance through keyword overlap and sentence length, used by \cite{agrawal-etal-2023-context}.

\noindent(3) \textbf{SBERT}~\cite{reimers-gurevych-2019-sentence}~is a semantic-based retrieval method, where we use ``paraphrase-mpnet-basev2'' following~\cite{li-etal-2023-unified}.

\noindent(4) \textbf{MMR}~\cite{ye-etal-2023-complementary} leverages BERTScore \cite{zhang2019bertscore} with maximal-marginal relevance for complementary example selection.

\noindent(5) \textbf{K-Means}~\cite{zhang2023automatic} performs $k$-means clustering to divide each dataset into four clusters. We then select examples randomly from each cluster.

\noindent(6) \textbf{BERT-FT}~\cite{devlin-etal-2019-bert} fine-tunes ``bert-base-uncased'' using the candidate pool examples.

\subsection{More Implementation Details}
\label{sec: settings}

Due to limited computational resources, test samples are restricted to 2,000 across the tasks: TwSenti, EmoC, P-Stance, and MNLI. Additionally, to accelerate inference, we load LLMs in fp16 precision. All experiments are conducted with an NVIDIA RTX A6000 GPU.

\section{Additional Results}
\label{sec: appendix more_results}

\subsection{Main Results in Accuracy}
\label{sec: more_acc}
For a more comprehensive comparison with the performance of baseline methods, we show additional main results in accuracy, as shown in Table~\ref{tab: main_acc}.  

\subsection{Ablation Results on More Baselines}
\label{more_ablation}
To comprehensively analyze the significance of each component within our framework, we conduct more ablation studies on two competitive baselines: Random and BM25. We report the results in Tables~\ref{tab:ablation_random} and~\ref{tab:ablation_bm25}, respectively.

\subsection{Effect on Subtle Sentiments for More Datasets}
\label{sec: more_effect}
\begin{figure}[!b]
    \centering
      \begin{subfigure}[b]{1\linewidth}
        \includegraphics[width=\linewidth]{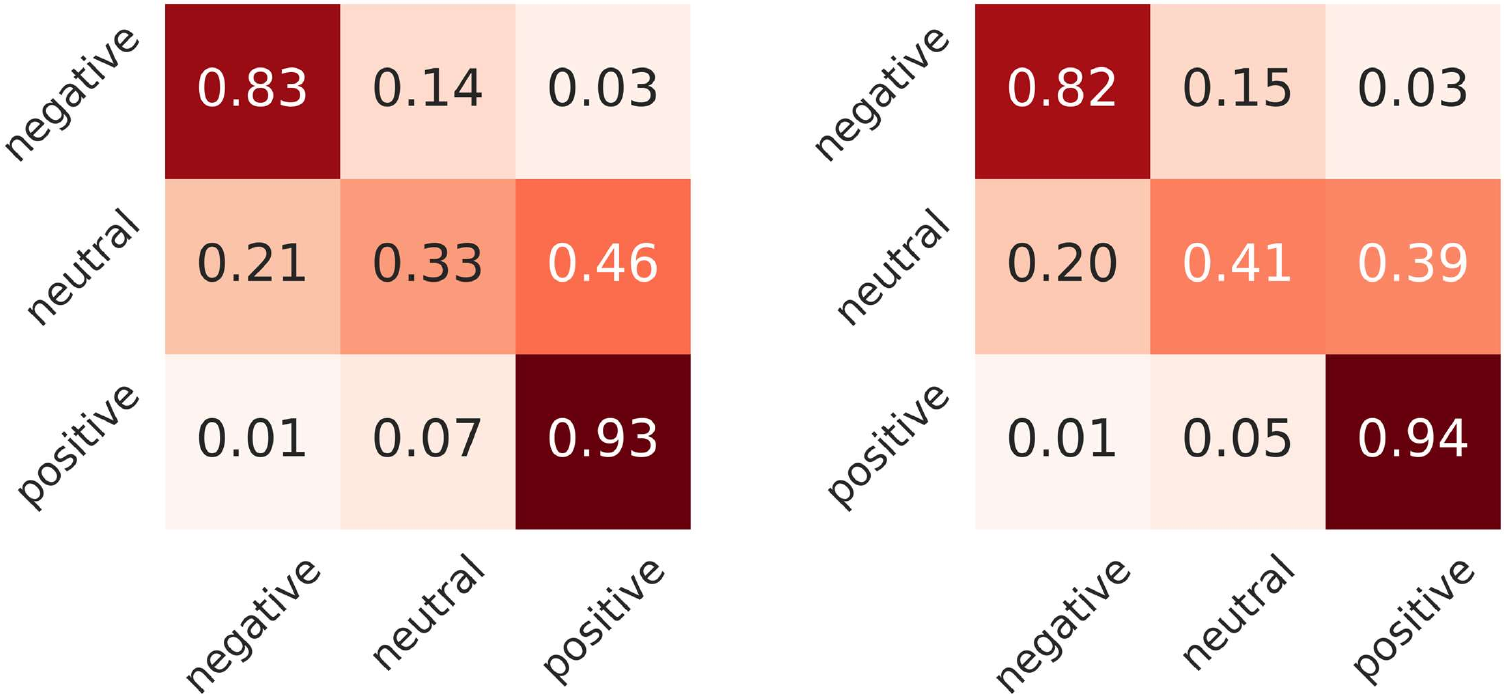}
        \caption{Normalized confusion matrices for the Rest dataset: BM25 (left) and BM25+Ours (right).}
      \end{subfigure}
      \begin{subfigure}[b]{1\linewidth}
        \includegraphics[width=\linewidth]{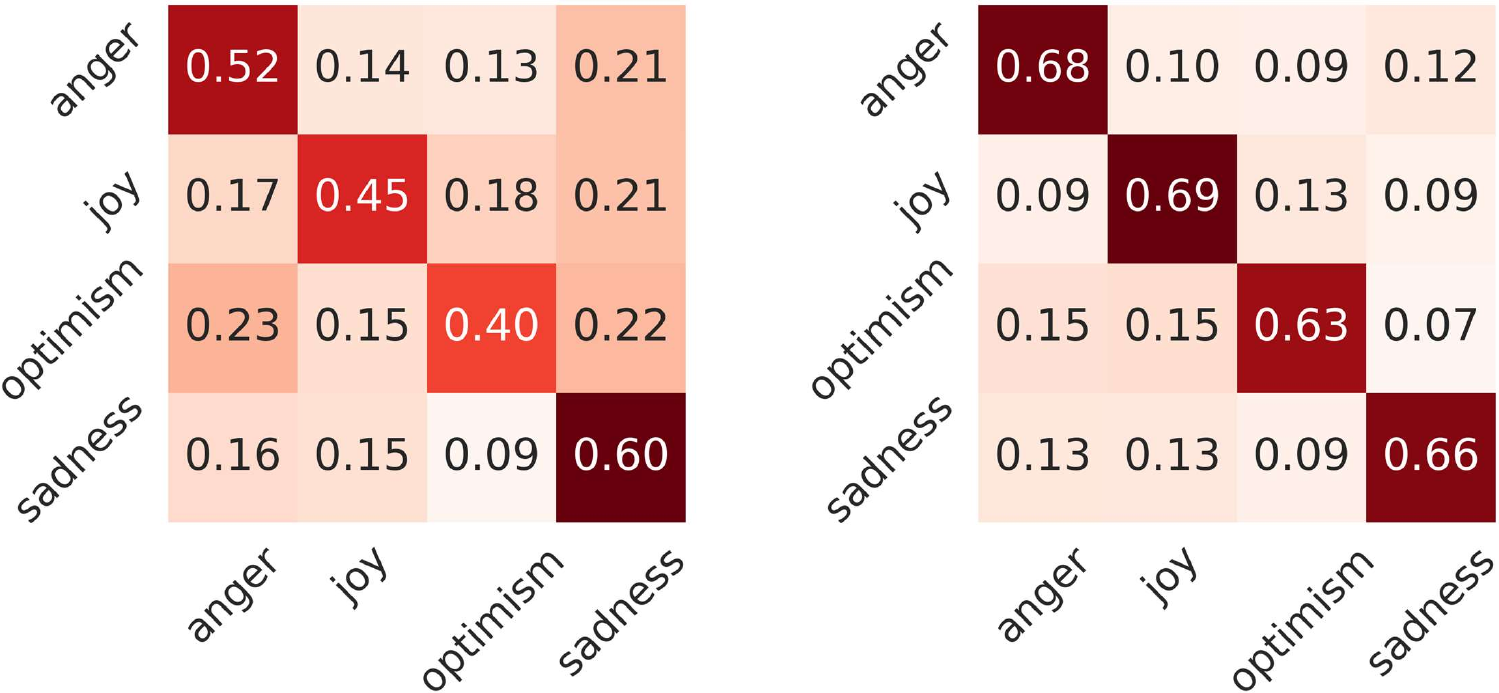}
        \caption{Normalized confusion matrices for the TwEmo dataset: BM25 (left) and BM25+Ours (right).}
      \end{subfigure}
      \begin{subfigure}[b]{1\linewidth}
        \includegraphics[width=\linewidth]{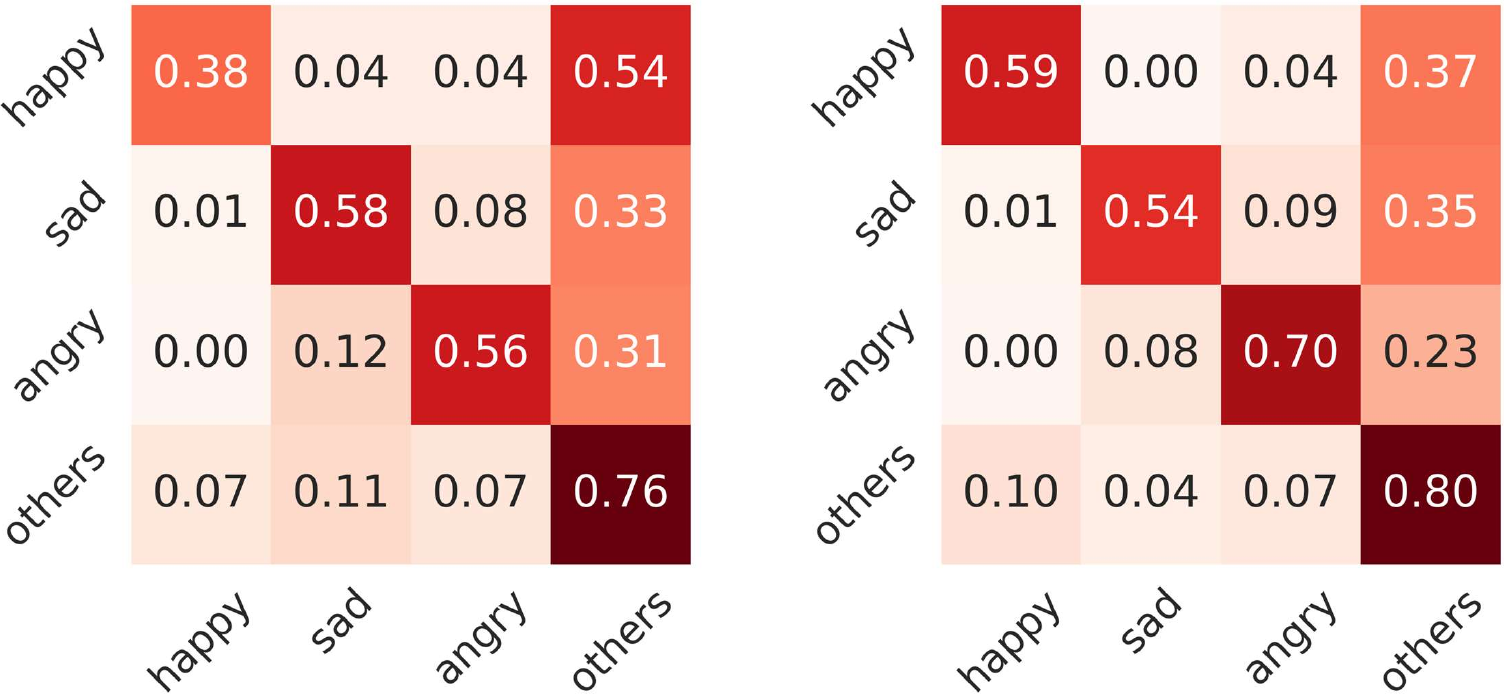}
        \caption{Normalized confusion matrices for the EmoC dataset: K-Means (left) and K-Means+Ours (right).}
      \end{subfigure}
  \caption{Effect on subtle sentiments for other datasets.}
  \label{fig:more_shift}
\end{figure}

To further illustrate how our framework corrects subtle sentiment understanding of the LLM and aligns predictions more closely with true labels, we visualize the improved prediction distributions on more datasets. The results are shown in Figure~\ref{fig:more_shift}.

\begin{table}[t]
  \centering
  \footnotesize
    \begin{tabular}{@{}cc|cc|ccc@{}}
    \toprule
    % \multicolumn{2}{c|}{ICL Input} & \multicolumn{2}{c|}{Ours} & \multicolumn{3}{c}{Dataset} \\
    % \midrule
    Inst & Label & Pred & Feed & Poem & Rest & TwEmo \\
    \addlinespace[0.15cm]
    % \midrule
    \cmark & \cmark & \cmark & \cmark & \textbf{64.37} & 71.16 & \textbf{60.91} \\
    \midrule
   % \cmark &  & \cmark & &  &  &  \\
   %  &  & \cmark & &  &  &  \\ \textcolor{orange} \textcolor{blue}
    \xmark & \cmark & \cmark & \cmark & 54.43 & \textbf{71.18} & 60.13 \\
    \cmark & \xmark & \cmark & \cmark & 52.93 & 67.45 & 51.63 \\
    \cmark & \cmark & \xmark & \cmark & 59.10 & 70.59 & 47.40 \\
    \cmark & \cmark & \textit{R} & \cmark & 63.35 & 69.21  & 60.66 \\
    \cmark & \cmark & \textit{Z} & \cmark & 64.08 & 68.58  & 60.76 \\
    \cmark & \cmark & \cmark & \xmark & 61.32 & 70.54 & 60.13  \\
    \bottomrule
  \end{tabular}
  \caption{Ablation study based on Random.}
  \label{tab:ablation_random}
\end{table}

\begin{table}[t]
  \centering
  \footnotesize
    \begin{tabular}{@{}cc|cc|ccc@{}}
    \toprule
    % \multicolumn{2}{c|}{ICL Input} & \multicolumn{2}{c|}{Ours} & \multicolumn{3}{c}{Dataset} \\
    % \midrule
    Inst & Label & Pred & Feed & Poem & Rest & TwEmo \\
    \addlinespace[0.15cm]
    % \midrule
    \cmark & \cmark & \cmark & \cmark & \textbf{61.27} & {\textbf{71.76}} & {\textbf{62.88}} \\
    \midrule
   % \cmark &  & \cmark & &  &  &  \\
   %  &  & \cmark & &  &  &  \\ \textcolor{orange} \textcolor{blue}
    \xmark & \cmark & \cmark & \cmark & 53.86 & 71.73 & 61.80 \\
    \cmark & \xmark & \cmark & \cmark & 50.42 & 67.74 & 51.95 \\
    \cmark & \cmark & \xmark & \cmark & 52.45 & 70.05 & 50.32 \\
    \cmark & \cmark & \textit{R} & \cmark & 60.07 & 70.18 & 62.64 \\
    \cmark & \cmark & \textit{Z} & \cmark & 61.09 & 69.82 & 62.02 \\
    \cmark & \cmark & \cmark & \xmark & 54.68 & 70.84 & 62.03 \\
    \bottomrule
  \end{tabular}
  \caption{Ablation study based on BM25.}
  \label{tab:ablation_bm25}
\end{table}

\subsection{More Case Studies}
\label{sec: more_case}
Additional cases are illustrated in Table~\ref{tab: more_case}. Observations reveal that incorporating prior predictions and feedback into examples not only corrects the sentiment labels but also yields more aligned explanations with human understandings.

\section{More Discussions}
\label{sec: appendix more_analysis}

\subsection{The Sensitivity of Feedback Prompt}
\label{sec: sensitivity}
To investigate the sensitivity of feedback design, we leverage the two hand-crafted feedback prompts shown in Table~\ref{tab: feedback_prompt} and experiment with four combinations. As presented in Table~\ref{tab:sen_feedback}, we find that on both datasets, the average variance in performance is within a 1\% margin. These results indicate that the performance is not highly sensitive to different feedback prompts.

\begin{table}[t]
    \centering
    \footnotesize
    \begin{tabular}{@{}C{1.35cm}C{1.15cm}C{1.15cm}C{1.15cm}C{1.15cm}}
        \toprule
        \multirow{2}{*}{Combination} & \multicolumn{2}{c}{\textbf{Rest}} & \multicolumn{2}{c}{\textbf{TwEmo}} \\
         \cmidrule(lr){2-3} \cmidrule(lr){4-5}
             & Acc & F1 & Acc & F1 \\
        \midrule
        FC\scalebox{1.0}[1.0]{-}1+FA\scalebox{1.0}[1.0]{-}1 & 81.83 & 71.76 & 66.92 & 62.88 \\
        \midrule
        FC\scalebox{1.0}[1.0]{-}1+FA\scalebox{1.0}[1.0]{-}2 & 82.31\textsubscript{+0.48} & 72.51\textsubscript{+0.75} & 67.00\textsubscript{+0.08} & 62.86\textsubscript{-0.02} \\

        FC\scalebox{1.0}[1.0]{-}2+FA\scalebox{1.0}[1.0]{-}1 & 82.22\textsubscript{+0.39} & 71.71\textsubscript{-0.05} & 68.30\textsubscript{+1.38} & 64.20\textsubscript{+1.32} \\

        FC\scalebox{1.0}[1.0]{-}2+FA\scalebox{1.0}[1.0]{-}1 & 82.10\textsubscript{+0.27} & 72.08\textsubscript{+0.32} & 67.98\textsubscript{+1.06} & 63.93\textsubscript{+1.05} \\
        \bottomrule
    \end{tabular}
    \caption{Results of different feedback (BM25+Ours).}
    \label{tab:sen_feedback}
\end{table}

\begin{table*}[!t]
    \centering
    \footnotesize
    \begin{tabular}{p{0.9cm}C{6.85cm}C{6.85cm}}
        \toprule
         & \textbf{Conventional In-context Learning} & \textbf{In-context Learning w/ Prediction Feedback (Ours)} \\
         
        \cmidrule(lr){2-2} \cmidrule(lr){3-3}
        
        Prompt & \textit{Task Instruction}: \ldots~~\textit{Examples}: $\{(x_i, y_i)\}$ & 
        \textit{Task Instruction}: \ldots~~\textit{Examples}: $\{(x_i, y'_i, y_i, fd)\}$ \\
        \midrule
        % \cmidrule(lr){2-3}
        Input & \multicolumn{2}{p{13.7cm}}{Another friend had to ask 3 times for parmesan cheese. What is the sentiment polarity of the aspect parmesan cheese ?} \\
        \cmidrule(lr){2-3}
        Output & negative (\wrongmark) & neutral (\correctmark) \\
        \cmidrule(lr){2-3}
        \multirow{8}{*}{Reason} & \multicolumn{1}{p{6.85cm}}{ 
        The sentence implies that the friend had to request the parmesan cheese multiple times, which suggests that \textbf{\textcolor{red}{the restaurant had issues with providing the cheese }}or that the staff was not attentive to the customer's requests. \textbf{\textcolor{red}{This negative experience with the parmesan cheese is evident in the sentence, and therefore, the sentiment polarity of the aspect "parmesan cheese" is negative.}} } & \multicolumn{1}{p{6.85cm}}{
       \textbf{\textcolor{deepgreen}{ The sentence does not convey any strong emotions, either positive or negative, towards the aspect "parmesan cheese".}} The fact that the friend had to ask three times for it suggests that there may have been some inconvenience or delay in receiving the cheese, \textbf{\textcolor{deepgreen}{but it does not convey any strong emotions or opinions about the cheese itself. Therefore, the sentiment polarity of the aspect "parmesan cheese" is neutral.}}}\\
        % \midrule
        % & \multicolumn{2}{c}{Case 2} \\
        \midrule
        % \cmidrule(lr){2-3}
        Input & \multicolumn{2}{p{13.7cm}}{my wife  did we all marry the same woman loudlycryingface} \\
        \cmidrule(lr){2-3}
        Output & angry (\wrongmark) & sad (\correctmark) \\
        \cmidrule(lr){2-2} \cmidrule(lr){3-3}
        \multirow{6}{*}{Reason} & \multicolumn{1}{p{6.85cm}}{ The sentence expresses a strong emotion, with the use of "loudly crying face" indicating that \textbf{\textcolor{red}{the speaker is upset or angry}} about something related to their wife. The sentence structure and word choice also suggest \textbf{\textcolor{red}{a sense of frustration or disbelief, which are common characteristics of angry emotion. }}} & \multicolumn{1}{p{6.85cm}}{ \textbf{\textcolor{deepgreen}{The sentence expresses a sense of disappointment and disbelief,}} as the speaker seems to be questioning how they all ended up married to the same woman. The use of the word "loudly crying face" also suggests that \textbf{\textcolor{deepgreen}{the speaker is emotional and possibly upset. The tone is introspective and melancholic, which is consistent with the emotion of sadness.}}}\\
        \bottomrule
    \end{tabular}
    \caption{More case studies of conventional ICL and our framework. }
    \label{tab: more_case}
\end{table*}

\subsection{Impact of the Number of Examples}
\begin{figure}[t]
    \centering    \includegraphics[width=0.95\linewidth]{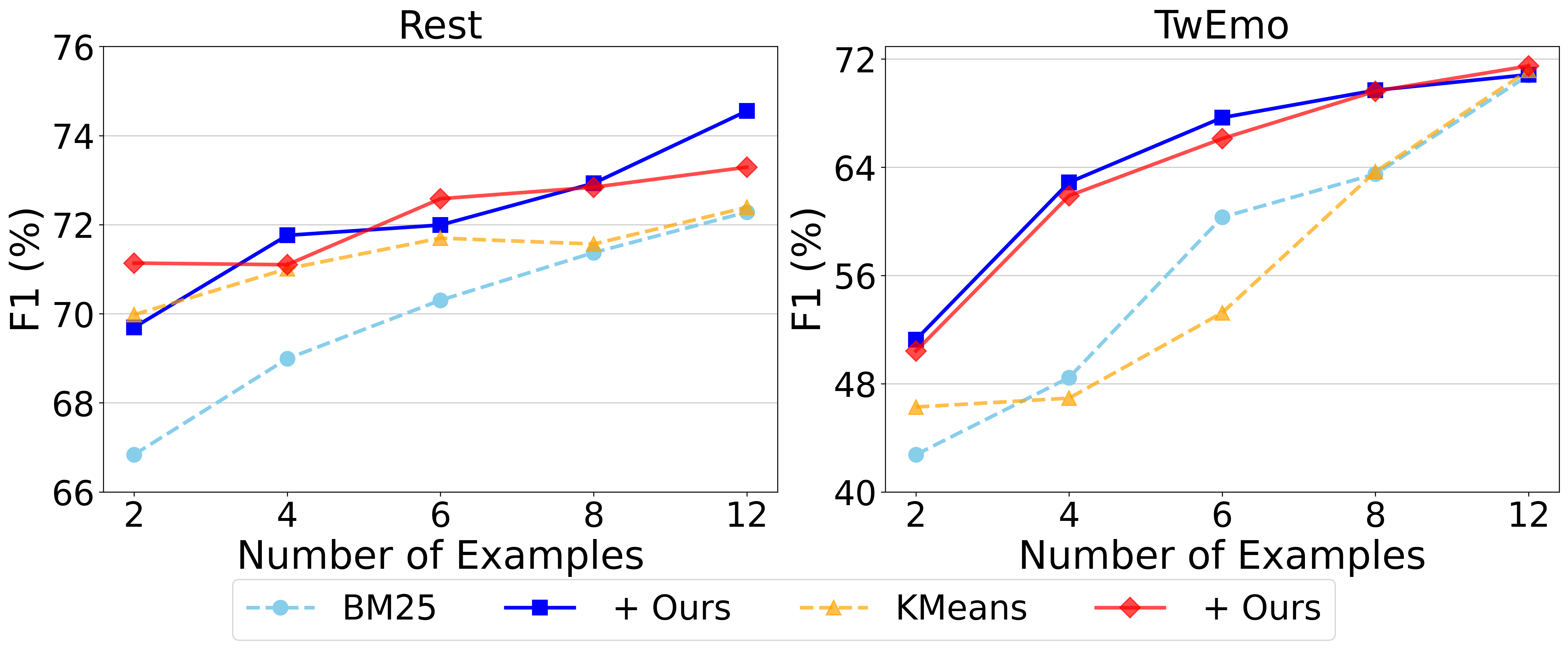}
    \caption{Effect of the quantity of in-context examples.}
    \label{fig: quantity}
\end{figure}

To analyze the effect of the example quantity, we perform experiments varying the number of examples ($k=2,4,6,8,12$), as depicted in Figure~\ref{fig: quantity}. First, we observe a consistent progressive trend of F1 as the number of examples increases. Second, our method generally yields significant performance gains on both datasets compared to the baselines, except for the 12-shot scenario on TwEmo. This suggests the proposed framework can positively influence the LLM to understand and analyze sentiment within an optimal context length.

\begin{table}[t]
    \centering
    \footnotesize
    \begin{tabular}{@{}p{1.5cm}C{0.8cm}C{0.8cm}C{0.8cm}C{0.8cm}C{0.8cm}@{}}
        \toprule
            \multicolumn{1}{c}{\multirow{2}{*}{Type}} & \multirow{2}{*}{Sort} & \multicolumn{2}{c}{\textbf{Rest}} & \multicolumn{2}{c}{\textbf{TwEmo}} \\
            \cmidrule(lr){3-4} \cmidrule(lr){5-6}
            & & Acc & F1 & Acc & F1 \\
            \midrule
            \multirow{2}{*}{Wrong First} & Desc\textsuperscript{\textdaggerdbl} & 81.83 & 71.76 & 66.92 & 62.88 \\
            & Asc & 81.35 & 71.12 & 65.94 & 62.18 \\
            \midrule
            \multirow{2}{*}{Correct First}& Desc & 81.77 & 70.04 & 66.97 & 62.57 \\
            & Asc & 82.08 & 70.77 & 65.47 & 60.90 \\
            \midrule
            \multirow{2}{*}{Alternating} & Desc & 81.74 & 71.05 & 66.71 & 63.11 \\
            & Asc & 82.10 & 70.83 & 66.60 & 62.11 \\
            \bottomrule
    \end{tabular}
    \caption{Effect of the order of examples (BM25+Ours). The standard setting is marked by \textdaggerdbl.}
    % Desc means descending order and Asc means ascending order.
    \label{tab: order}
\end{table}

\subsection{Impact of the Order of Examples}
To investigate the impact of example ordering, we first categorize three strategies: prioritizing wrong examples (wrong first), prioritizing correct examples (correct first), and alternating between the two. We then apply both ascending and descending arrangements based on retrieval scores. On this basis, we experiment with five extra permutations, as shown in Table~\ref{tab: order}. We find that the performance of Rest remains stable regardless of permutations, with a minor standard deviation of 0.51 in F1. Conversely, on TwEmo, descending ordering generally outperforms ascending ones.
These findings suggest that although our framework is robust against the variability of strategy, the consideration of specific arrangement methods could be important.

\begin{table*}[!t]
    \centering
    \footnotesize
    \begin{tabular}{cC{1.2cm}ccccC{1cm}c}
    
        \toprule
        Task & Dataset & Domain & Train & Dev & Test & Classes & Labels \\
        \midrule
        %\multicolumn{7}{c}{\textit{Sentiment Analysis Tasks}} \\
        %\midrule
         \multirow{4}{*}{SC} 
          & SST-2 & Movie Reviews & 6,920 & 872 & 1,821 & 2 & positive, negative \\
          & TwSenti & Social Media & 45,615 & 2,000 & 12,284 & 3 & positive, negative, neutral \\
          & Poem & Literature & 843 & 105 & 104 & 3 & positive, negative, neutral \\
          & Finance & Financial & 1,358 & 453 & 453 & 3 & positive, negative, neutral \\
         \midrule
         \multirow{3}{*}{ASC} 
          & Rest & Customer Reviews & 3,608 & 454 & 1,119 & 3 & positive, negative, neutral \\
          & Laptop & Customer Reviews & 2,282 & 283 & 682 & 3 & positive, negative, neutral \\
          & Twitter & Social Media & 6,248 & - & 692 & 3 & positive, negative, neutral \\
         \midrule
         \multirow{2}{*}{ED} 
          & EmoC & Social Media & 30,160 & - & 5,509 & 4 & happy, sad, angry, others \\
          & TwEmo & Social Media & 3,257 & 374 & 1,421 & 4 & anger, joy, optimism, sadness \\
         \midrule
        %  \multicolumn{7}{c}{\textit{Other Tasks}} \\
         \midrule
         Stance & P\scalebox{1.0}[1.0]{-}Stance & Social Media & 17,756 & 2,282 & 2,207 & 2 & favor, against \\
         
         Irony & TwIrony & Social Media & 2,862 & 955 & 784 & 2 & irony, non-irony \\
         
         NLI & MNLI & General & 263,789 & 3,000 & 9,796 & 3 & entailment, contradiction, neutral \\
         \bottomrule
    \end{tabular}
    \caption{The statistics of investigated datasets.}
    \label{tab: dataset}
\end{table*}

\begin{table*}[h]
  \centering
  \footnotesize
    \begin{tabular} {p{1.2cm}C{1.14cm}C{1.14cm}C{1.14cm}C{1.14cm}C{1.14cm}C{1.14cm}C{1.14cm}C{1.14cm}C{1.14cm}}
    \toprule
    \multirow{2}{*}{Method} & \multicolumn{4}{c}{\textbf{\footnotesize{Sentiment Classification}}} & \multicolumn{3}{c}{\textbf{\footnotesize{Aspect Sentiment Classification}}} & \multicolumn{2}{c}{\textbf{\footnotesize{Emotion Detection}}} \\
    \cmidrule(lr){2-5} \cmidrule(lr){6-8} \cmidrule(lr){9-10}
    & \multicolumn{1}{c}{\texttt{SST-2}} & \multicolumn{1}{c}{\texttt{TwSenti}} & \multicolumn{1}{c}{\texttt{Poem}} & \multicolumn{1}{c}{\texttt{Finance}} & \multicolumn{1}{c}{\texttt{Rest}} & \multicolumn{1}{c}{\texttt{Laptop}} & \multicolumn{1}{c}{\texttt{Twitter}} & \multicolumn{1}{c}{\texttt{EmoC}} & \multicolumn{1}{c}{\texttt{TwEmo}} \\ 
    \midrule
     BERT\scalebox{1.0}[1.0]{-}FT\textsuperscript{\textdagger} & 85.02 & 54.87 & 75.64 & 90.80 & 73.19 & 74.37 & 56.31 & 65.12 & 65.87 \\
    \midrule
    Random & 89.64 & 55.13 & 55.77  & 75.28 & 79.80 & 77.27 & 54.00 & 69.05 & 52.99 \\
    ~~~+~Ours & \cellcolor[gray]{0.9}{91.40\textsubscript{+1.76}} & \cellcolor[gray]{0.9}{60.37\textsubscript{+5.24}} & \cellcolor[gray]{0.9}{69.23\textsubscript{+13.46}} & \cellcolor[gray]{0.9}{78.44\textsubscript{+3.16}} & \cellcolor[gray]{0.9}{81.23\textsubscript{+1.43}} & 77.27\textsubscript{+0.00} & \cellcolor[gray]{0.9}{56.74\textsubscript{+2.74}} & \cellcolor[gray]{0.9}{74.97\textsubscript{+5.92}} & \cellcolor[gray]{0.9}{66.26\textsubscript{+13.27}} \\
    \midrule
    
    BM25 & 90.06 & 55.15 & 49.68 & 53.13 & 80.25 & 75.05 & 50.05 & 71.23 & 52.69 \\
    ~~~+~Ours & \cellcolor[gray]{0.9}{91.70\textsubscript{+1.64}} & \cellcolor[gray]{0.9}{59.20\textsubscript{+4.05}} & \cellcolor[gray]{0.9}{66.03\textsubscript{+16.35}} & \cellcolor[gray]{0.9}{64.46\textsubscript{+11.33}} & \cellcolor[gray]{0.9}{81.83\textsubscript{+1.58}} & \cellcolor[gray]{0.9}{76.11\textsubscript{+1.06}} & \cellcolor[gray]{0.9}{55.20\textsubscript{+5.15}} & \cellcolor[gray]{0.9}{75.95\textsubscript{+4.72}} & \cellcolor[gray]{0.9}{66.92\textsubscript{+14.23}} \\
    \midrule
    
    SBERT & 87.65 & 49.98 & 47.76 & 44.44 & 78.08 & 69.57 & 49.76 & 72.33 & 50.88 \\
    ~~~+~Ours & \cellcolor[gray]{0.9}{91.32\textsubscript{+3.67}} & \cellcolor[gray]{0.9}{55.57\textsubscript{+5.59}} & \cellcolor[gray]{0.9}{62.82\textsubscript{+15.06}} & \cellcolor[gray]{0.9}{54.67\textsubscript{+10.23}}  & \cellcolor[gray]{0.9}{80.28\textsubscript{+2.20}} & \cellcolor[gray]{0.9}{74.31\textsubscript{+4.74}} & \cellcolor[gray]{0.9}{55.30\textsubscript{+5.54}} & \cellcolor[gray]{0.9}{78.37\textsubscript{+6.04}} & \cellcolor[gray]{0.9}{64.34\textsubscript{+13.46}} \\
    \midrule
    MMR & 89.45 & 50.55 & 50.00 & 51.29 & 78.37 & 71.04 & 50.00 & 70.48 & 53.51 \\
    ~~~+~Ours & \cellcolor[gray]{0.9}{92.49\textsubscript{+3.04}} & \cellcolor[gray]{0.9}{56.98\textsubscript{+6.43}} & \cellcolor[gray]{0.9}{67.95\textsubscript{+17.95}} & \cellcolor[gray]{0.9}{56.36\textsubscript{+5.07}}  & \cellcolor[gray]{0.9}{79.95\textsubscript{+1.58}} & \cellcolor[gray]{0.9}{74.00\textsubscript{+2.96}} & \cellcolor[gray]{0.9}{54.53\textsubscript{+4.53}} & \cellcolor[gray]{0.9}{76.28\textsubscript{+5.80}} &  \cellcolor[gray]{0.9}{65.63\textsubscript{+12.12}} \\
    \midrule
     K\scalebox{1.0}[1.0]{-}Means & 88.65 & 56.12 & 50.96 & 75.86 & 81.47 & 77.85 & 54.38 & 72.48 & 52.69 \\
    ~~~+~Ours & \cellcolor[gray]{0.9}{92.09\textsubscript{+3.44}} & \cellcolor[gray]{0.9}{61.37\textsubscript{+5.25}} & \cellcolor[gray]{0.9}{72.44\textsubscript{+21.48}} & \cellcolor[gray]{0.9}{78.15\textsubscript{+2.29}} & 81.23\textsubscript{-0.24} & 77.69\textsubscript{-0.16} & \cellcolor[gray]{0.9}{56.84\textsubscript{+2.46}} & \cellcolor[gray]{0.9}{76.55\textsubscript{+4.07}} & \cellcolor[gray]{0.9}{66.96\textsubscript{+14.27}} \\
    \bottomrule
  \end{tabular}
  \caption{Main results in Acc\%. Fine-tuning methods are marked by \textdagger. }
  \label{tab: main_acc}
\end{table*}

\begin{table*}[h]
    \centering
    \footnotesize
    \begin{tabular}{c|ll}
        \toprule
            \multicolumn{1}{c}{ }& \multicolumn{1}{c}{Feedback on correct examples } & \multicolumn{1}{c}{Feedback on wrong examples}  \\
            \midrule
            ~~~FC-1~~~ & You are correct! & You are wrong! \\
            ~~~FA-1~~~ & Make sure your prediction is accurate. ~~~~~~~& Stay determined and keep moving forward. \\
            \midrule
            ~~~FC-2~~~ & The answer is accurate. & The answer is incorrect. \\
            ~~~FA-2~~~ & Please keep up the good work. & Please adjust to ensure the prediction is correct.~~~ \\
            \bottomrule
    \end{tabular}
    \caption{Different feedback prompts. FC is feedback on correctness and FA is feedback for analysis.}
    \label{tab: feedback_prompt}
\end{table*}

\begin{table*}
    \centering
        \begin{tabular}{C{1.4cm}|p{13.5cm}}
            \toprule
            Task & \multicolumn{1}{c}{In-context Learning Prompts} \\
            \midrule
            % \multicolumn{2}{c}{\textit{Sentiment Analysis}} \\
            % \midrule
            SC & \begin{minipage}{13.5cm}
                    \textbf{Instruction:} Recognize the sentiment of the sentence. Here are some examples: \\ 
                    \textbf{Examples:} \ldots \\
                    \phantom{\textbf{Examples:}} Sentence: \textit{text} $x_i$\\
                    \phantom{\textbf{Examples:}} Prediction: \textit{prior prediction} $y'_i$\\
                    \phantom{\textbf{Examples:}} Correct Label: \textit{label} $y_i$ \\
                    \phantom{\textbf{Examples:}} \textit{feedback on correct (or wrong) examples} \\
                    \phantom{\textbf{Examples:}} \ldots \\
                    \textbf{Test Input:} Sentence: \textit{text} $x$ \\
                    \phantom{\textbf{Test Input:}} Correct Label: 
                \end{minipage} \\
            \midrule
            ASC & \begin{minipage}{13.5cm}
                    \textbf{Instruction:} Recognize the sentiment polarity for the given aspect term in the sentence. Here are some examples: \\ 
                    \textbf{Examples:} \ldots \\
                    \phantom{\textbf{Examples:}} Sentence: \textit{text} $x_i$ What is the sentiment polarity of the aspect \textit{aspect} ?\\
                    \phantom{\textbf{Examples:}} Prediction: \textit{prior prediction} $y'_i$\\
                    \phantom{\textbf{Examples:}} Correct Label: \textit{label} $y_i$ \\
                    \phantom{\textbf{Examples:}} \textit{feedback on correct (or wrong) examples} \\
                    \phantom{\textbf{Examples:}} \ldots \\
                    \textbf{Test Input:} Sentence: \textit{text} $x$ What is the sentiment polarity of the aspect \textit{aspect} ?\\
                    \phantom{\textbf{Test Input:}} Correct Label: 
                \end{minipage} \\
            \midrule
            ED & \begin{minipage}{13.5cm}
                    \textbf{Instruction:} Recognize the emotion of the sentence. Here are some examples: \\ 
                    \textbf{Examples:} Same as SC \\
                    \textbf{Test Input:} Sentence: \textit{text} $x$ \\
                    \phantom{\textbf{Test Input:}} Correct Label: 
                \end{minipage} \\
            \midrule
            %\multicolumn{2}{c}{\textit{Task Generalization}} \\ 
            %\midrule
            ~\newline~\newline~\newline~\newline Stance\newline Detection \newline~\newline~\newline~\newline~ & \begin{minipage}[t]{13.5cm}
                    \textbf{Instruction:} Recognize the stance of the sentence to the given target. Here are some examples: \\ 
                    \textbf{Examples:} \ldots \\
                    \phantom{\textbf{Examples:}} Sentence: \textit{text} $x_i$ What is the attitude of sentence toward target \textit{target} ?\\
                    \phantom{\textbf{Examples:}} Prediction: \textit{prior prediction} $y'_i$\\
                    \phantom{\textbf{Examples:}} Correct Label: \textit{label} $y_i$ \\
                    \phantom{\textbf{Examples:}} \textit{feedback on correct (or wrong) examples} \\
                    \phantom{\textbf{Examples:}} \ldots \\
                    \textbf{Test Input:} Sentence: \textit{text} $x$ What is the attitude of sentence toward target \textit{target} ?\\
                    \phantom{\textbf{Test Input:}} Correct Label: 
                \end{minipage} \\
            \midrule
            ~\newline Irony\newline Detection\newline ~ & \begin{minipage}[t]{13.5cm}
                    \textbf{Instruction:} Determine whether the sentence is ironic or not. Here are some examples: \\ 
                    \textbf{Examples:} Same as SC\\
                    \textbf{Test Input:} Sentence: \textit{text} $x$ \\
                    \phantom{\textbf{Test Input:}} Correct Label: 
                \end{minipage}\\
            \midrule
            NLI & \begin{minipage}{13.5cm}
                    \textbf{Instruction:} Recognize textual entailment between the 2 texts. Here are some examples: \\ 
                    \textbf{Examples:} \ldots \\
                    \phantom{\textbf{Examples:}} Premise: \textit{text1} $x_{i1}$\\
                    \phantom{\textbf{Examples:}} Hypothesis: \textit{text2} $x_{i2}$\\
                    \phantom{\textbf{Examples:}} Prediction: \textit{prior prediction} $y'_i$\\
                    \phantom{\textbf{Examples:}} Correct Label: \textit{label} $y_i$ \\
                    \phantom{\textbf{Examples:}} \textit{feedback on correct (or wrong) examples} \\
                    \phantom{\textbf{Examples:}} \ldots \\
                    \textbf{Test Input:} Premise: \textit{text1} $x_{test1}$ \\
                    \phantom{\textbf{Test Input:}} Hypothesis: \textit{text2} $x_{test2}$ \\
                    \phantom{\textbf{Test Input:}} Correct Label: 
                \end{minipage}\\
            \bottomrule
        \end{tabular}
        \caption{The prompt and format of in-context learning for each task.}
        \label{tab: prompt}
\end{table*}

\end{document}